%% file: main.tex
\documentclass[10pt,twocolumn,letterpaper]{article}

\usepackage[pagenumbers]{cvpr} 

\usepackage{graphicx}
\usepackage{amsmath}
\usepackage{amssymb}
\usepackage{booktabs}

\input{custom}

\usepackage{algorithm}

\usepackage[pagebackref,breaklinks,colorlinks]{hyperref}

\usepackage[capitalize]{cleveref}
\crefname{section}{Sec.}{Secs.}
\Crefname{section}{Section}{Sections}
\Crefname{table}{Table}{Tables}
\crefname{table}{Tab.}{Tabs.}

\def\cvprPaperID{2164} 
\def\confName{CVPR}
\def\confYear{2023}

\begin{document}

\title{Fair Robust Active Learning by Joint Inconsistency}

\author{
Tsung-Han Wu$^1$ \qquad Hung-Ting Su$^1$ \qquad Shang-Tse Chen$^1$ \qquad Winston H. Hsu$^{1,2}$
\\
\\
$^1$National Taiwan University \qquad $^2$Mobile Drive Technology
}

\maketitle

\begin{abstract}

\input{0_abstract.tex}
\end{abstract}

\section{Introduction}
\input{1_introduction_new.tex}

\section {Related Work}
\input{2_relatedworks.tex}

\section {Method}

\input{3_method_new.tex}

\section {Experiments}

\input{4_experiments.tex}

\section {Conclusion}
\input{5_conclusion.tex}

\clearpage

\section*{Acknowledgement}
This work was supported in part by the National Science and Technology Council, under Grant MOST 110-2634-F-002-051 and MOST 110-2222-E-002-014-MY3, as well as Mobile Drive Technology Co., Ltd (MobileDrive). We are grateful to the National Center for High-performance Computing. We also thank Hsin-Ying Lee and Jhih-Ciang Wu for the kind suggestions on figures and paper writing.


{\small
\bibliographystyle{ieee_fullname}
\bibliography{ref.bib}
}

\clearpage

\appendix
\begin{center}
\large{\textbf{Supplementary Material}}
\end{center}
\input{supp.tex}

\end{document}

%% file: custom.tex
\usepackage{multirow}
\usepackage{xcolor}
\usepackage{algpseudocode}
\usepackage{setspace}
\usepackage{amsfonts}
\usepackage{colortbl}
\definecolor{gray}{rgb}{0.5,0.5,0.5} 
\definecolor{frenchblue}{rgb}{0.0, 0.45, 0.73}
\definecolor{gray}{rgb}{0.5,0.5,0.5} 
\definecolor{green}{rgb}{0, 0.4, 0} 
\definecolor{orange}{rgb}{1, 0.5, 0} 	
\definecolor{mahogany}{rgb}{0.75, 0.25, 0.0}
\definecolor{purple}{rgb}{0.6, 0, 0.6}
\definecolor{darkgreen}{rgb}{0, 0.4, 0.4} 
\definecolor{red}{rgb}{1.0, 0, 0}
\definecolor{plotpurple}{rgb}{0.2353, 0.2, 0.90196}
\definecolor{plotorange}{rgb}{1.0, 0.6, 0.2}
\definecolor{plotgreen}{rgb}{0.2, 0.784313, 0.2}
\definecolor{plotred}{rgb}{1.0, 0.2, 0.392}
\definecolor{LightCyan}{rgb}{0.88,1,1}

\newcommand{\std}[1]{\footnotesize{$\pm$#1} }

\def\supp{supplementary material}

%% file: 0_abstract.tex
Fairness and robustness play vital roles in trustworthy machine learning. Observing safety-critical needs in various annotation-expensive vision applications, we introduce a novel learning framework, \textbf{Fair Robust Active Learning (FRAL)}, generalizing conventional active learning to fair and adversarial robust scenarios. This framework allows us to achieve standard and robust minimax fairness with limited acquired labels. In FRAL, we then observe existing fairness-aware data selection strategies suffer from either ineffectiveness under severe data imbalance or inefficiency due to huge computations of adversarial training. To address these two problems, we develop a novel Joint INconsistency (JIN) method exploiting prediction inconsistencies between benign and adversarial inputs as well as between standard and robust models. These two inconsistencies can be used to identify potential fairness gains and data imbalance mitigations. Thus, by performing label acquisition with our inconsistency-based ranking metrics, we can alleviate the class imbalance issue and enhance minimax fairness with limited computation. Extensive experiments on diverse datasets and sensitive groups demonstrate that our method obtains the best results in standard and robust fairness under white-box PGD attacks compared with existing active data selection baselines. 

%% file: 1_Introduction_new.tex
\label{sec:intro}

While supervised deep learning methods have achieved remarkable success in a variety of computer vision tasks, the cost of labeling a large amount of data required for such a training paradigm is a huge burden.
As a result, some utilize \textit{active learning} (AL) techniques to achieve high performance by gradually selecting limited but valuable data for manual labeling \cite{ijcnn,wang2016cost, gal2016dropout,gal2017deep,ash2020babdge,kirsch2019batchbald,sener2018coreset}.

Recently, in addition to reaching high performance, \textit{fairness} and \textit{robustness} have played increasingly vital roles in trustworthy visual applications. For example, a facial attribute recognition model is commonly used in biometric systems to protect the safety and confidentiality of individuals. To ensure the safety and fairness of such systems, the model should never exhibit low performance against specific genders with or without adversarial attacks. Nonetheless, no existing work has explored the possibility of achieving this under limited annotations.


\begin{figure*}[!t]
    \centering
    \includegraphics[width=\linewidth]{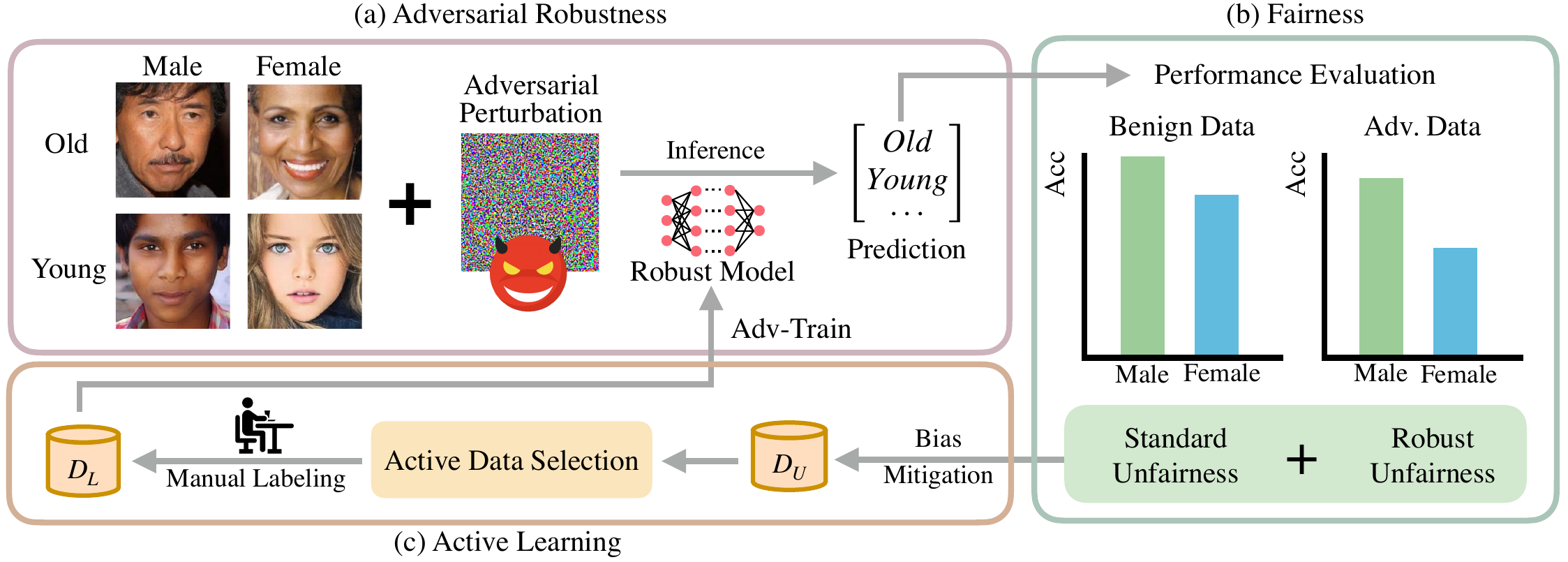}
    \caption{\textbf{A facial attribute recognition system based on our FRAL framework.} (a) In this system, a robust classifier predicting the young or the old is first adversarially trained with the current labeled dataset $D_L$ against adversarial attacks. (b) However, the model may display \textbf{standard unfairness} on benign samples or \textbf{robust unfairness} on adversarially attacked images between males and females. (c) Consequently, we iteratively acquire manual annotations of limited samples in the unlabeled pool $D_U$ using a specific active data selection strategy. After being finetuned with these newly annotated data, the classifier can achieve better results in the underperforming (female) group, thus increasing the standard and robust minimax fairness (return to (a)). With the FRAL framework, we can alleviate performance discrimination with limited manual annotation effort in standard and adversarial robust settings.} 
    \label{fig:fig1}
    \vspace{-0.7em}
\end{figure*}

Observing this, we introduce a novel learning framework, \textbf{Fair Robust Active Learning (FRAL)}, generalizing conventional AL to fair and robust scenarios. With this redesigned framework, we are able to reach \textit{standard} and \textit{robust} fairness with a few acquired labels. To elaborate, in both standard and adversarial robust settings, we can achieve the widely used minimax fairness, i.e., the highest worst-case performance across groups \cite{diana2021minimax}. Fig. \ref{fig:fig1} illustrates the benefits of the FRAL framework for annotation-expensive applications with safety-critical requirements.



Under the FRAL framework, due to the need to satisfy robustness constraints, existing fairness-aware active data selection methods would face two challenges. First, because performance disparities between classes are amplified under adversarial training \cite{xu2021robust, wang2021imbalanced}, existing methods \cite{shekhar2021adaptive, abernethy2020active} may perform poorly on low-frequency classes of disadvantaged groups, resulting in low fairness. Second, as computational costs grow dramatically under adversarial training \cite{NEURIPS2019_7503cfac}, some other approaches \cite{anahideh2022fair,sharaf2022promoting} may suffer from expensive computations of measuring the expected fairness gain for each sample in the unlabeled dataset.
To conquer these two problems, we propose a method called Joint INconsistency (JIN). 
Specifically, on benign data, disparities between classes and predicted errors increase from standard to robust models \cite{wang2021imbalanced,xu2021robust,tsipras2018robustness}. Similarly, for the robust model, incorrect outputs and class imbalance performance grow from benign to adversarial inputs \cite{wang2021imbalanced,xu2021robust,cisse2017parseval,zhang2019theoretically}. Hence, inspired by prior inconsistency-based active selection methods \cite{wang2012inconsistency,yu2022consistency}, we leverage the two prediction disagreements to identify potential performance improvements and label imbalance mitigation. In practice, we estimate the worst-performing group and then select the top-ranked samples in it for labeling at each active learning round. Unlike randomly drawing samples from the worst group \cite{shekhar2021adaptive, abernethy2020active}, we select more data in minor classes and thus work better under severe label imbalance. Also, by maintaining a standard-trained model on a small labeled set instead of measuring the values of all unlabeled data by adversarial training \cite{anahideh2022fair,sharaf2022promoting}, we become computationally efficient.

We validate the efficacy of our method under white-box PGD attacks \cite{madry2018towards} in a wide variety of visual applications, including facial attribute recognition \cite{zhang2017utkface}, object classification \cite{darlow2018cinic}, and cell type identification \cite{tschandl2018ham10000}. Besides, we utilize several sensitive attributes in our experiments, including age, gender, and membership in a group. In various combinations, our JIN method outperforms existing active data selection approaches in standard and robust fairness metrics. Ablation studies prove the effectiveness of all our proposed components. Also, extensive experiments on different model backbones and multiple sensitive groups further demonstrate our generalization ability. To sum up, our contributions are listed as follows: 
\begin{itemize}
    \setlength\itemsep{0em}
    \item We introduce a new learning framework, Fair Robust Active Learning (FRAL), practical for annotation-expensive applications with safety-critical needs.
    \item Under the FRAL framework, we design JIN, a novel data selection strategy, to solve the computation and class imbalance issue of prior fairness-aware methods.
     \item Our method surpasses existing active data selection baselines in both standard and robust fairness metrics under different experimental settings and datasets.
\end{itemize}

%% file: 2_relatedworks.tex
\smallskip\noindent{\bf Fairness in ML.} Fairness is a fundamental problem in the field of ML. Many prior methods have pointed out that biases across sensitive groups are widely presented in ML models and datasets \cite{schnabel2016recommendations,lambrecht2019algorithmic,raji2019actionable,karkkainen2021fairface}. Also, several debiasing training strategies \cite{menon2018cost,oneto2019taking,ustun2019fairness} 
are proposed to achieve fairness from different aspects, including making predictions independent of sensitive features \cite{calders2010three,dwork2012fairness}, yielding equal prediction odds on favored results \cite{hardt2016equality}, or maximizing the accuracy for disadvantaged groups \cite{martinez2020minimax}. Recently, some practices discussed fairness under adversarial attacked scenarios or via actively collecting a small amount of data. We elaborate on these methods in the followings.

\vspace{0.5em}

\smallskip\noindent{\bf Adversarial Robustness.}
Research on adversarial robustness can be roughly divided into attack and defense. Adversarial attacks aim to generate adversarial samples misclassified by ML models by adding the least perturbations to benign data, while defensive approaches seek to enhance model robustness against such attacks. Common adversarial scenarios are black-box and white-box threat models based on knowing all or nothing of the victim's ML model.

In the past few years, many classical attacks, such as FGSM \cite{goodfellow2014explaining} and PGD \cite{madry2018towards}, produced adversarial examples by back-propagating loss functions. On the other hand, defensive methods utilized obfuscating gradients \cite{athalye2018obfuscated} or training with robust optimization \cite{madry2018towards,zhang2019trades,zhang2021_GAIRAT} against adversaries. 

Recently, few approaches considered the intersection of adversarial robustness and fairness. Some discussed the bias between classes in adversarial training and proposed a training framework to mitigate this issue \cite{xu2021robust}; Others analyzed differences in robustness to adversarial samples between sensitive groups and developed a simple regularization method to address the problem \cite{nanda2021fairness}.


Parallel to prior work of designing training algorithms, we delve into ways to achieve equalized performance and equalized robustness between groups via active data collection, which is more suitable for many real-world applications where manual labels are difficult to obtain.



\vspace{0.5em}

\smallskip\noindent{\bf Active Learning for Fairness.} Conventional active learning aims to attain high model performance by actively querying limited manual annotations. Common label acquisition strategies can be roughly divided into two types: model uncertainty and data diversity. In the past, uncertainty-based methods \cite{ijcnn,wang2016cost, gal2016dropout,gal2017deep} collected data with the least model confidence for manual labeling, aiming to reduce model uncertainty after appending these data into training. Recently, some methods \cite{ash2020babdge,kirsch2019batchbald,sener2018coreset} increased the data diversity within a query batch, further improving the labeling efficiency.

More recently, several studies utilized active learning techniques to achieve fairness. Some analyzed the efficacy of existing uncertainty-based active sampling methods under fairness evaluations \cite{branchaud2021can}. Others designed fairness-aware data selection strategies by estimating expected unfairness reduction \cite{anahideh2022fair} or utilizing meta-learning \cite{sharaf2022promoting}. Still others developed adaptive sampling policies specifically for fairness with theoretical foundations \cite{shekhar2021adaptive, abernethy2020active}.  However, these works merely focused on fairness without robust settings.

To the best of our knowledge, we first introduce the task targets not only equalized performance on benign data but also equalized robustness to adversarial attacks. In this task, those fairness-centric methods cannot afford the computations of adversarial training \cite{anahideh2022fair,sharaf2022promoting} or struggle with fairness improvement under severe label imbalance \cite{shekhar2021adaptive, abernethy2020active}. To this end, we propose an active selection method based on the properties of adversarial training, which is fundamentally different from all existing methods.

%% file: 3_method_new.tex
\subsection{Problem Definition}
\label{sec:problem_def}

In this work, we introduce Fair Robust Active Learning (FRAL), a novel learning framework to reach equalized standard and robust performance between sensitive groups by actively acquiring limited labeled data.
Specifically, let $x$ and $y$ represent an input sample and a target label; $z \in Z$ indicates a sensitive attribute or membership in a group. Taking the scenario in Fig. \ref{fig:fig1} as an example, $x$ refers to a face image, $y$ is a class in $Y = \{\textrm{Young}, \textrm{Old}\}$, and $z$ is an attribute belonging to $Z = \{\textrm{Male}, \textrm{Female}\}$. Assume we have a training set $D$ composed of a small labeled set $D_L = \{(x_i, y_i, z_i)\}_{i=1}^N$ and another unlabeled pool $D_U = \{(x_j, z_j)\}_{j=1}^M$. In our framework, we first train a robust deep learning model $M_R: x \rightarrow y$ with $D_L$ by adversarial training. Then, once we observe performance discrimination in standard or adversarial settings, we select a few samples in $D_U$ with active selection strategies for manual labeling and further training.

To prevent achieving fairness by deliberately degrading the performance of dominant groups, we utilize the widely-used minimax fairness \cite{diana2021minimax} rather than predictive disparities between groups as the major objective for the problem. In other words, all methods are required to achieve fairness by maximizing the standard performance and adversarial robustness of the least favorable group. Formally, given a white-box adversarial attack function $\mathcal{A}(x, \epsilon) \rightarrow \tilde{x}$ ($\epsilon$ is a pre-defined maximum perturbation range), a testing set $D_{T}$, and a robust deep learning model $M_R$, the standard fairness $\mathcal{F}_{std}$ and the robust fairness $\mathcal{F}_{rob}$ are defined as the probability of correct predictions in the worst group as follows:
\begin{equation}
\begin{aligned}
& \mathcal{F}_{std} = \mathop{\min}\limits_{z \in Z}\ \mathbb{P}\{M_R(x) = y\ |\ (x, y, z) \in D^z_{T} \}, \\
& \mathcal{F}_{rob} = \mathop{\min}\limits_{z \in Z}\ \mathbb{P}\{M_R(\mathcal{A}(x, \epsilon)) = y\ |\ (x, y, z) \in D^z_{T} \},
\end{aligned}
\end{equation}
where $D^z_{T}$ is a subset of $D_{T}$ with the same attribute $z$. 

\algnewcommand\algorithmicforeach{\textbf{for each}}
\algdef{S}[FOR]{ForEach}[1]{\algorithmicforeach\ #1\ \algorithmicdo}

\begin{algorithm}[!h]
    \begin{algorithmic}[1]
    \setstretch{1.2}
    \State \textbf{Input:} training set $D = \{D_L, D_U\}$, validation set $D_V$, adversarial attacks $\mathcal{A}$ with a perturbation range $\epsilon$, maximum active rounds $K$, and labeling budgets $B$.
    \State \textbf{Output:} An adversarial robust model $M_{R}$.
    \State \textbf{Init:} $M_R \leftarrow {\textrm{Adv-TRAIN}(D_L, \epsilon)}$
    \For{$k \leftarrow$ $1$ \textbf{to} $K$}
    \State $z^* \leftarrow \textrm{EVAL}(M_R, \mathcal{A}, \epsilon, D_V)$ \Comment{Get the worst group}
    \State $X \leftarrow \{x \ |\ (x, z) \in D_U\ \wedge \ z = z^*\}$
    \State {$I \leftarrow$ Get inconsistency scores for $X$ via Eq. 2, 3, 4}
    \State $X^* \leftarrow \{x\ |\ x \in X \wedge I_x \textrm{ is of the top-} B \textrm{ values}\}$
    
    
    \State $Y^* \leftarrow \textrm{Manual-Labeling}(X^*)$
    \State $D_L \leftarrow D_L\ \bigcup\ \{(x, y, z^*)\ |\ x \in X^*, y \in Y^*\}$
    \State $D_U \leftarrow D_U \setminus \{(x, z^*)\ |\ x \in X^*\}$
    \State $M_R \leftarrow{\text{Adv-FINETUNE}(D_L, \epsilon)}$ 
    \EndFor
    \State \Return $M_R$
    \caption{Fair Robust Active Learning by JIN}
    \label{alg:debias}
    \end{algorithmic}
\end{algorithm}

\vspace{-1em}

\subsection{Overview}

We propose a novel method called Joint INconsistency (JIN), general for various model architectures and adversarial training strategies under the FRAL framework. Algo. \ref{alg:debias} shows the complete algorithm of our method, which consists of 3 main steps: (1) Model Initialization: Train the robust model $M_R$ with the initial labeled set $D_L$ by adversarial training. (2) Joint Inconsistency Sample Ranking: For each active learning round, estimate the worst-performing group and obtain joint inconsistency scores for all samples belonging to that group (Sec. \ref{sec:jin}). (3) Label Acquisition: Select top-ranked samples for manual labeling until running out of budgets, update $D_L$ along with $D_U$ accordingly, and fine-tune $M_R$ to boost fairness objectives  (Sec. \ref{sec:label}). 

\subsection{Joint Inconsistency Sample Ranking}
\label{sec:jin}

As stated in Sec. \ref{sec:intro}, prior fairness-aware selection methods either randomly sample data in the worst group \cite{shekhar2021adaptive, abernethy2020active} or estimate expected fairness gain via meta-learning \cite{sharaf2022promoting} or fine-tuning on all unlabeled data \cite{anahideh2022fair} for label acquisition. In adversarial training scenarios, however, the former suffer from severe data imbalance problems and the latter are confronted with an overwhelming computational burden. As a result, we design an efficient and effective sample ranking method via joint inconsistency to identify valuable samples for labeling, which is detailed below.


\vspace{0.5em}

{\smallskip\noindent{\bf Worst group estimation.}} A simple way to improve the standard and robust minimax fairness, \ie, $\mathcal{F}_{std}$ and $\mathcal{F}_{rob}$, is to select more valuable data within the worst-performing group for manual labeling and further training. To achieve the goal, we first estimate the worst-performing group $z^*$ with the validation set at the beginning of each active selection round\footnote{In real-world applications, it is easy to identify the least favorable group by analyzing user feedback or data. Detailed discussions and additional experiments on this issue are provided in Sec. \ref{sec:discussion}.}. Then, we select a few samples to annotate based on our designed inconsistency scores, which will be discussed in the following sections.
%


\vspace{0.5em}

\smallskip\noindent{\bf Inconsistency for standard fairness.} 
To maximize the expected standard fairness gain from a few acquired labels, we select the samples with the highest inconsistency score between the robust model $M_R$ and an auxiliary standard-trained model $M_S$. Inspired by prior active learning practices leveraging disagreeing prediction as data selection criteria \cite{wang2012inconsistency,yu2022consistency}, we hypothesize that the disagreement between $M_R$ and $M_S$ indicates the potential knowledge gain from annotation.
Specifically, assuming we have $M_S$ and $M_R$, the standard inconsistency score $I^{std}_{x}$ for a sample $x$ in the unlabeled set is defined as the prediction disagreement between $M_S$ and $M_R$ on benign data:
\begin{equation}
    I^{std}_{x} = D_{\text{KL}}(p(x, M_S)\ ||\ p(x, M_R)),
    \label{eq:std}
\end{equation}
where $p(x, M)$ indicates the predicted probability distribution of sample $x$ from model $M$ and $D_{\text{KL}}(\cdot||\cdot)$ means KL-divergence between the two distributions.

\begin{figure}[t]
    \centering
    \includegraphics[width=\linewidth]{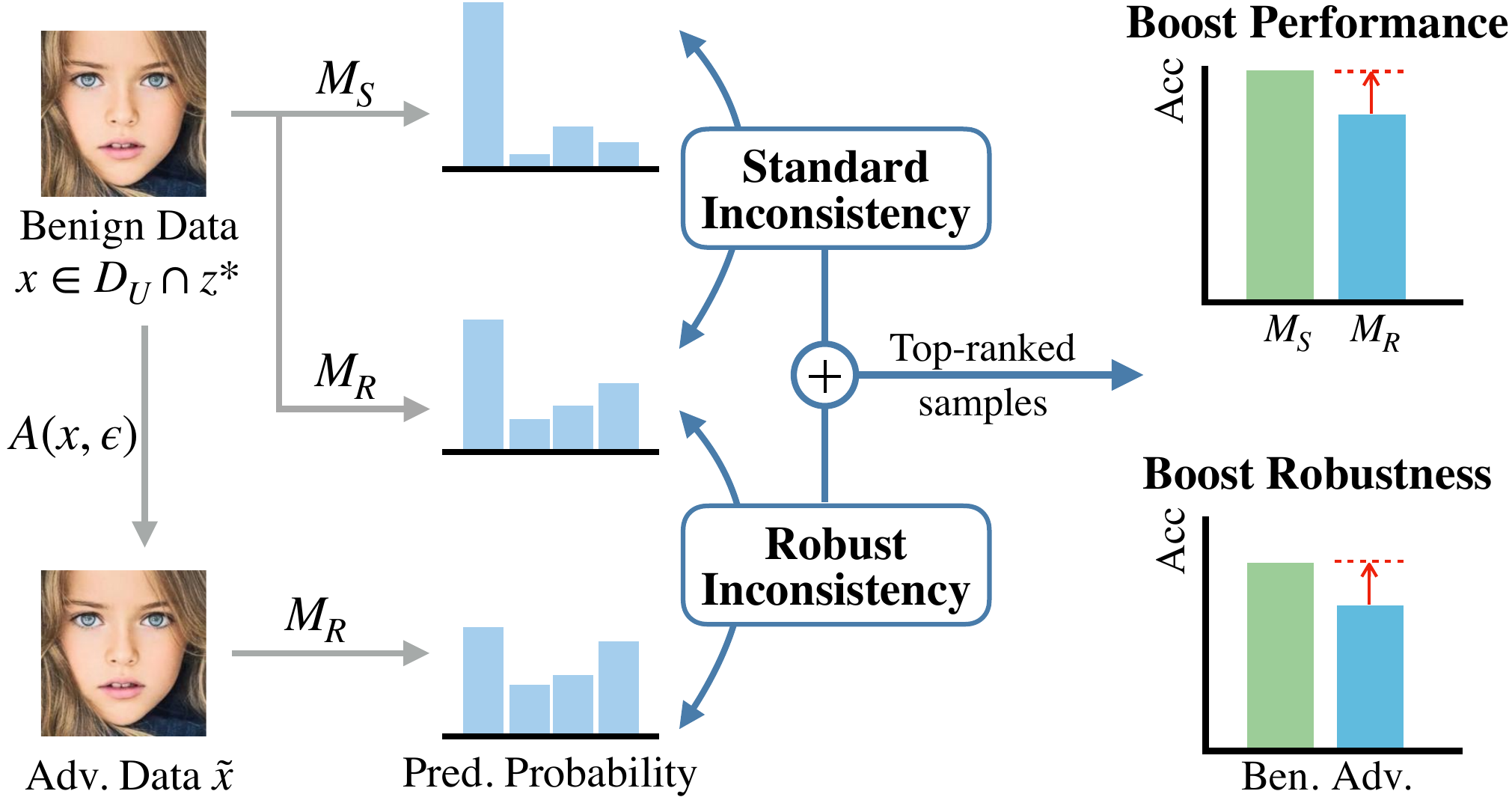}
    \caption{\textbf{The diagram of our proposed Joint INconsistency (JIN) method.} The standard inconsistency metric measures the divergence between the predicted probability distribution of benign data from the standard model $M_S$ and the robust model $M_R$. Similarly, robust inconsistency estimates the divergence of two distributions between benign and adversarial samples from the robust model $M_R$. By acquiring labels of limited samples belonging to the worst-performing group with top-ranked inconsistency scores, the standard performance and adversarial robustness of the group can be enhanced, mitigating the unfairness issue.}
    \label{fig:fig2}
    \vspace{-1em}
\end{figure}

The motivation of Eq. \ref{eq:std} comes from theories in adversarial training. \cite{tsipras2018robustness} proves that improving the robustness of ML models would sacrifice performance on benign data, and \cite{wang2021imbalanced,xu2021robust} observe that the issue of class-imbalanced performance on benign data becomes more severe under adversarial training. Based on these two studies, $M_S$ has better performance and a milder class-imbalance problem than $M_R$ on benign samples. Therefore, the performance drop of $M_R$ could be alleviated by acquiring benign samples with utmost inconsistency between $M_S$ and $M_R$. Observing this, we calculate the inconsistency score with KL divergence between output distributions of two models and use it to  measure the potential performance gain of the sample. By selecting top-ranked samples from the most unfavorable group $z^*$, we improve the standard minimax fairness $\mathcal{F}_{std}$ and alleviate the class imbalance problem. 
Our implementation is illustrated at the top of Fig. \ref{fig:fig2}. For each active selection step, we first maintain an auxiliary model $M_S$ by standard training with a small labeled set $D_L$. Then, all unlabeled benign samples are fed to $M_S$ and $M_R$ to obtain two different predicted probabilities. Lastly, we calculate the standard inconsistency scores $I^{std}$ for all samples. The scores will be used in the label acquisition process, which we will cover later.







\vspace{0.5em}

{\smallskip\noindent{\bf Inconsistency for robust fairness.}} Similar to standard inconsistency, another inconsistency between benign and adversarial samples output by the same $M_R$ is utilized to measure the expected robust fairness improvement.
Formally, the robust inconsistency score $I^{rob}_{x}$ of a sample $x$ in the unlabeled set is defined as follows:
\begin{equation}
    I^{rob}_{x} = D_{\text{KL}}(p(x, M_R)\ ||\ p(\mathcal{A}(x, \epsilon)), M_R),
    \label{eq:rob}
\end{equation}
where $\mathcal{A}(x, \epsilon)$ is a white-box adversarial attack function identical to the definition in Sec. \ref{sec:problem_def}.

The intuition of Eq. \ref{eq:rob} also stems from the properties of adversarial robustness. To begin with, model smoothness is considered to be highly correlated with adversarial robustness \cite{cisse2017parseval,zhang2019theoretically}. Moreover, under adversarial training, robust models generally exhibit larger performance disparities between classes given adversarial inputs than benign inputs \cite{wang2021imbalanced,xu2021robust}. 
Consequently, regardless of the data imbalance issue, the adversarial sensitive samples could be easily identified by measuring the benign and adversarial outputs from $M_R$ and utilized to boost robustness.
Specifically, in each active selection round, we use the KL distance of the two prediction distributions $I^{rob}$ as another indicator to rank the samples from the worst group to enhance $\mathcal{F}_{rob}$. This process is illustrated at the bottom of Fig. \ref{fig:fig2}.



So far, we obtain two critical indicators, $I^{std}$ and $I^{rob}$, that can identify potential standard and robust fairness improvement. As our inconsistency metrics can address performance disparities between classes, we excel on severely class-imbalanced datasets compared to \cite{shekhar2021adaptive, abernethy2020active}. Besides, our method requires only multiple model inferences and low-cost standard training on a small labeled set rather than expensive adversarial finetuning on the unlabeled set. Therefore, it is much more computationally efficient than \cite{sharaf2022promoting,anahideh2022fair}.



\subsection{Label Acquisition}
\label{sec:label}

As described above, the metrics $I^{std}$ and $I^{rob}$ are used to select samples that can boost the standard and robust fairness respectively. Thus, to simultaneously maximize $\mathcal{F}_{std}$ and $\mathcal{F}_{rob}$ with the least manual annotations, we simply take the sum of these two scores as our final inconsistency metric for active data selection as follows:
\begin{equation}
    I_{x} = N(I^{std}_{x}) + N(I^{rob}_{x}),
\end{equation}
where $N(\cdot)$ is a standardization function that turns the value into an average of 0 and a standard deviation of 1.



After obtaining the score $I$ for all samples belonging to the worst-performing group, we acquire the labels of top-ranked samples until running out of labeling budgets. Next, we append these labeled data into $D_L$ and remove them from $D_U$. Finally, we fine-tune the robust model $M_R$ with the updated $D_L$ by adversarial training and proceed to the next active learning round.

%% file: 4_experiments.tex
\subsection{Experimental Settings}

We describe datasets, evaluation metrics, training protocols, and baselines in this section. 

\vspace{0.5em}

\noindent {\textbf{Datasets.}} We use three different datasets in our experiments: UTKFace~\cite{zhang2017utkface}, CINIC-10~\cite{darlow2018cinic} and HAM-10000~\cite{tschandl2018ham10000}. For the UTKFace facial attribute dataset, we construct two sensitive groups (young and old) and perform the 4-race (White, Black, Asian, and Indian) classification task. For the CINIC-10 dataset, integrated by CIFAR-10 and partially synthesized ImageNet, we classify ten objects and treat membership in the two domains as sensitive attributes. For the HAM-10000 skin lesion dataset, we recognize seven cell types and use genders as sensitive attributes. 

We directly use the official CINIC-10 data split in our experiments and construct the other two datasets ourselves due to the lack of such information. For the UTKFace, we first filter out face photos belonging to the ``Other" race and put an age-related attribute tag on each photo based on the provided age metadata. Specifically, we evenly divide all faces into two groups (Young and Old) with a threshold of 30 years old. For the HAM-10000 dataset, we simply filter out repeated images and samples without a corresponding sex attribute. After the above preprocessing steps, we randomly split the two datasets into training and validation sets with a ratio of 7:3. We measure the effectiveness of all methods on the validation set.

\vspace{0.5em}

\noindent {\textbf{Evaluation Metrics.}} 
We utilize the commonly used minimax fairness, \ie, highest worst group performance, as the primary fairness evaluation metric \cite{martinez2020minimax}. For a fair comparison with prior methods, we include prediction disparities between the highest and lowest performance groups as another fairness criterion \cite{verma2018fairness}. Also, we report the average group performance to see whether a serious fairness performance trade-off exists.
For the UTKFace and CINIC-10 datasets, we use average accuracy for performance evaluation. As for the HAM-10000 dataset, we adopt the F1-score as the performance evaluation metric following prior work \cite{sandler2018mobilenetv2} on account of a severe label imbalance problem.

\vspace{0.5em}

\noindent {\textbf{Training Protocol.}} For all experiments, we utilize MobileNetV2 \cite{sandler2018mobilenetv2} as the network backbone owing to the great performance on adversarial training with high training and inference speed. For the adversarial robust scenario, we set the threat model as white-box PGD-5 attacks with maximum perturbation range $\ell_{\infty}=4/255$ and step size $\alpha=2/255$. In the model initialization and finetuning stage, we use the TRADES loss \cite{zhang2019trades} to train our robust model. Note that in order to conquer serious label imbalance, we apply random oversampling when training models with HAM-10000 dataset. More training details and computing infrastructure are reported in the \supp{}.

As for the active learning setting, we perform five active data selection rounds ($K=5$). For all three datasets, we randomly divide the training set $D$ into 20\% initial labeled set $D_L$ plus 80\% unlabeled set $D_U$ as initialization. Then, the labeling budget $B$ for each round is 2\% of $|D|$.


\vspace{0.5em}

\input{tables/big_tab_new.tex}

\noindent {\textbf{Baselines.}} We compare our designed JIN method with eight active data selection baselines. For active learning baselines, in addition to random selection (RAND), we include three diverse and representative methods, including an uncertainty-based method (ENT \cite{ijcnn}), a diversity approach (CSET \cite{sener2018coreset}), and a hybrid strategy (BADGE \cite{ash2020babdge}). Besides, we utilize four fairness-aware data selection baselines. They involve FairAL \cite{anahideh2022fair}, a method leveraging expected fairness gain, and three adaptive sampling methods, including naive worst-group random selection (G-RAND), MinMax \cite{martinez2020minimax}, and OPT \cite{shekhar2021adaptive}. Note that as FairAL focuses on demographic parity rather than accuracy parity, we modify its source code to fit into our minimax fairness setting for fair comparisons. We do not include PANDA \cite{sharaf2022promoting} in baselines due to unaffordable computations of meta-learning.

\subsection{Main Results}

Tab. \ref{tab:main_result} and Tab. \ref{tab:main_result2} compare the effectiveness of various methods on three datasets. On the UTKFace and CINIC-10 datasets, the robust model favors the old group and samples belonging to CIFAR-10, respectively. On the HAM-10000 dataset, the robust model does not necessarily favor males or females over several active learning rounds. Our proposed JIN method achieves the highest minimax fairness on all three datasets. Besides, in most cases, we deliver the lowest predictive disparity without degradation in average standard performance and adversarial robustness. 



\vspace{0.5em}

\noindent {\textbf{Performance comparison to active learning baselines.}}  In most cases, our JIN method outperforms RAND, CSET, and BADGE by more than one standard deviation in two different fairness metrics with or without adversarial attacks. We observe that ENT obtains better standard fairness than three other active learning methods, which is identical to previous related research \cite{branchaud2021can}. Still, our method reaches better standard fairness than ENT. Under the adversarial robust setting, our method achieves significant fairness advantages over ENT, including higher minimax fairness and lower predictive disparities. This indicates that in addition to selecting hard samples to boost standard fairness similar to ENT, our method can further identify adversarial sensitive samples in the worst-performing group for labeling.





\input{tables/big_tab_new2.tex}
\input{tables/label_imbl.tex}

\noindent {\textbf{Performance comparison to fairness-aware selection.}} Our method has significant performance advantages over group-aware adaptive sampling strategies (G-RAND, MinMax, OPT), including a lead of more than two standard deviations on the standard minimax fairness and a lead of nearly one standard deviation on the robust minimax fairness. Furthermore, compared to the original label-balanced dataset CINIC-10, our method achieves more benefits on two other datasets with label imbalance.

To investigate this issue, we draw the correlation between label distribution and per-class performance in Tab. \ref{tab:ham} with the HAM-10000 dataset. As our JIN method acquires more rare class samples belonging to the worst group, like ``mel" (Melanoma) and ``akiec" (Actinic keratoses), we obtain higher F1-score in these classes compared to all existing baselines, especially G-RAND. In contrast, for the dominant class, ``nv" (melanocytic nevi), though selecting relatively few samples, our method still achieves comparable results. These analyses demonstrate that under the FRAL framework, existing adaptive sampling strategies cannot deal with datasets with uneven label distribution, but our proposed JIN method greatly improves this situation.





We compare our method with FairAL, which is also a method for estimating the expected fairness increase per sample. In robust minimax fairness, our JIN method surpasses FairAL by more than one standard deviation. As for standard fairness, our method is still superior to FairAL by more than $1\%$ minimax fairness on the CINIC-10 and HAM-10000 datasets and achieves remarkably lower predictive disparity on the UTKFace dataset. In terms of performance, the huge gain in robust fairness proves that our method can select more adversarially sensitive samples for labeling than FairAL. Besides, in terms of computational efficiency, FairAL requires a lot more time than us, which will be explained as follows.


\input{tables/computation}

\noindent {\textbf{Comparison on computational costs.}}
We report the computation burdens of different active data selection methods in Tab. \ref{tab:computation}. Compared to ENT and G-RAND requiring low computations, our method takes more time because of maintaining a standard-trained auxiliary model. With this help, our method achieves significantly better standard and robust fairness than theirs as shown in Tab. \ref{tab:main_result} and Tab. \ref{tab:main_result2}. 

While FairAL takes the longest time among the four methods, its fairness performance is still inferior to ours. On the CINIC-10 dataset, FairAL even takes more than twice as much initial adversarial training for data selection, proving this approach is impractical for real-world applications. The main reason is that CINIC-10 contains much more samples than the other two datasets. Thus, FairAL requires more adversarial finetuning to estimate the potential fairness.

To conclude, under the FRAL framework, our JIN data selection method achieves a great trade-off between fairness performance and computational costs. Among all active data selection baselines, we achieve the best standard and robust fairness using fewer than 30\% of the initial adversarial training computations.







\subsection{Discussions}
\label{sec:discussion}

In addition to the main comparison in Tab. \ref{tab:main_result}, we conduct in-depth experiments on the UTKFace dataset. Below we first verify the effectiveness of the proposed components. Then, we show that our method is applicable to different deep neural networks and multiple sensitive groups.


\vspace{0.5em}

\noindent {\textbf{Ablation studies.}} Tab. \ref{tab:abl_result} shows the efficacy of our proposed standard and robust inconsistency metrics. As shown in a comparison of the first and the second row, the standard inconsistency enhances more standard minimax fairness along with standard group average scores; while the robust inconsistency achieves better results in two adversarial robust metrics. In the third row, where the two metrics are used together, we observe a large improvement in average robustness as well as robust fairness, but only a small drop in standard performance. This result suggests that jointly using two metrics may be the optimal strategy to achieve both equalized performance and equalized robustness under the FRAL framework.

\input{tables/abl}




\noindent {\textbf{Efficacy of known worst group.}} Because traditional active learning methods, unlike ours, do not know the worst group when selecting data for labeling, we specifically discuss the issue of knowing this information. To begin with, getting information on the worst group is a reasonable setting as it is easy to identify bias in real-world applications \cite{buolamwini2018gender}. Additionally, recent related work \cite{xu2021robust} also uses a validation set to obtain information on each group as we do.


We also conduct additional experiments for a fair comparison. We extend ENT, the best active learning method, to the G-ENT active selection strategy. G-ENT executes the same data selection algorithm as ENT, except that it only samples data from the worst group, which is the same setting as JIN. As shown in Tab. \ref{tab:abl_result_main}, in terms of fairness and average score, G-ENT performs worse than ours, even worse than the original ENT. It proves that simply performing traditional active learning from the worst group does not yield better results. We infer the reason might be sampling bias.

\input{tables/worst_group.tex}

\noindent {\textbf{Generalization on various model architectures.}} We conduct experiments using ResNet18 model architecture \cite{he2016deep} under the same training protocol. As shown in Tab. \ref{tab:gaware}, our method surpasses all representative baselines in standard and robust minimax fairness. Furthermore, identical to Tab. \ref{tab:main_result}, our method is able to select more adversarial sensitive samples from the worst-performing group than FairAL, thus achieving better robust minimax fairness. Overall, the results confirm that our JIN method is applicable to various neural network backbones.


\input{tables/gaware}

\vspace{0.5em}

\noindent {\textbf{Generalization on multiple sensitive groups.}} To validate the efficacy of various active selection methods on non-binary sensitive attributes, we further conduct experiments on the UTKFace gender prediction task and treat the four different races as sensitive groups (White, Black, Asian, and Indian). We use the same adversarial training protocol mentioned before with five active data selection rounds. The only difference is that we only randomly choose 10\% $D$ to initialize $D_L$ and set merely 1\% $|D|$ as the labeling budget for each round. The reason for using less labeled data is that the gender prediction task is simpler than the 4-race prediction. As shown in Tab. \ref{tab:multiple_sens}, we outperform all baselines in both standard and robust fairness. The result demonstrates the generalization ability of our method.

\input{tables/multiple_sens}


%% file: tables/big_tab_new.tex
\begin{table*}[t]
\centering
\setlength\tabcolsep{3pt}
\resizebox{1.0\linewidth}{!}{
\begin{tabular}{c|ccc|ccc||ccc|ccc}
\toprule
&\multicolumn{6}{c||}{UTKFace 4-Race Classification (sensitive groups: \{Young, Old\})}&\multicolumn{6}{c}{CINIC-10 Classification (sensitive groups: \{CIFAR-10, ImageNet\})}\\
\midrule
\multicolumn{1}{c|}{
\multirow{2}{*}{\begin{tabular}[c]{@{}c@{}}Methods\end{tabular}}}&\multicolumn{3}{c|}{Standard Accuracy (\%)}&\multicolumn{3}{c||}{Robust Accuracy (\%)}&\multicolumn{3}{c|}{Standard Accuracy (\%)}&\multicolumn{3}{c}{Robust Accuracy (\%)}\\
&Worst ($\uparrow$)&Disp ($\downarrow$) &Avg ($\uparrow$)&Worst ($\uparrow$)&Disp ($\downarrow$)&Avg ($\uparrow$)&Worst ($\uparrow$)&Disp ($\downarrow$) &Avg ($\uparrow$)&Worst ($\uparrow$)&Disp ($\downarrow$)&Avg ($\uparrow$)\\

\midrule

Init. AT&67.58\std{0.30}&5.38\std{0.25}&70.27\std{0.31}&52.98\std{0.08}&7.26\std{0.31}&56.61\std{0.06}& 52.53\std{0.17}&12.48\std{0.21}&58.77\std{0.40}&31.29\std{0.11}&10.64\std{0.23}&36.61\std{0.03}\\

\midrule
    RAND&70.57\std{0.21}&4.32\std{0.03}&72.73\std{0.21}&55.63\std{0.06}&7.71\std{0.02}&59.49\std{0.07}&55.53\std{0.53}&12.14\std{0.55}&61.60\std{0.61}&37.01\std{0.43}&11.43\std{0.37}&42.73\std{0.60}\\
    ENT&74.10\std{0.79}&2.45\std{0.48}&75.33\std{0.56}&56.94\std{0.64}&6.60\std{0.33}&60.25\std{0.56}&56.23\std{0.52}&11.30\std{0.39}&61.88\std{0.64}&36.29\std{0.40}&10.52\std{0.42}&41.55\std{0.51}\\
    CSET&71.44\std{0.46}&3.47\std{0.52}&73.31\std{0.21}&56.55\std{0.19}&6.42\std{0.49}&59.76\std{0.05}&55.28\std{0.44}&12.94\std{0.51}&61.75\std{0.52}&36.73\std{0.27}&12.22\std{0.52}&\textbf{42.74\std{0.39}}\\
    BADGE&72.63\std{0.20}&3.53\std{0.23}&74.31\std{0.13}&56.94\std{0.40}&6.07\std{0.20}&59.98\std{0.50}&55.86\std{0.38}&11.96\std{0.44}&61.84\std{0.37}&36.66\std{0.30}&11.04\std{0.38}&42.18\std{0.29}\\
    \midrule
    G-RAND&72.37\std{0.32}&2.15\std{0.26}&73.45\std{0.23}&56.60\std{0.04}&6.07\std{0.33}&59.63\std{0.13}&55.56\std{0.43}&\textbf{10.76\std{0.61}}&60.94\std{0.66}&36.71\std{0.35}&10.02\std{0.41}&41.72\std{0.59}\\
    MinMax&71.35\std{0.24}&3.27\std{0.28}&72.98\std{0.20}&56.95\std{0.22}&6.59\std{0.12}&60.25\std{0.21}&55.52\std{0.49}&11.32\std{0.63}&61.22\std{0.60}&36.69\std{0.46}&10.52\std{0.53}&41.95\std{0.47}\\
    OPT&71.99\std{0.31}&2.76\std{0.23}&73.37\std{0.20}&57.09\std{0.33}&6.11\std{0.19}&60.15\std{0.24}&55.78\std{0.33}&10.90\std{0.37}&61.23\std{0.49}&36.90\std{0.29}&~~9.96\std{0.36}&41.88\std{0.50}\\
    FairAL&74.74\std{0.31}&2.20\std{0.13}&\textbf{75.84\std{0.25}}&56.94\std{0.16}&6.64\std{0.17}&\textbf{60.47\std{0.07}}&56.35\std{0.45}&10.98\std{0.44}&61.84\std{0.58}&36.25\std{0.29}&10.40\std{0.33}&41.45\std{0.37}\\
    \midrule
    \cellcolor{LightCyan}\textbf{JIN}&\cellcolor{LightCyan}\textbf{75.07\std{0.53}}&\cellcolor{LightCyan}\textbf{1.35\std{0.09}}&\cellcolor{LightCyan}75.74\std{0.49}&\cellcolor{LightCyan}\textbf{57.39\std{0.10}}&\cellcolor{LightCyan}\textbf{5.69\std{0.30}}&\cellcolor{LightCyan}60.10\std{0.25}&\cellcolor{LightCyan}\textbf{57.37\std{0.67}}&\cellcolor{LightCyan}11.16\std{0.52}&\cellcolor{LightCyan}\textbf{62.95\std{0.68}}&\cellcolor{LightCyan}\textbf{37.10\std{0.45}}&\cellcolor{LightCyan}\textbf{~~9.84\std{0.45}}&\cellcolor{LightCyan}42.02\std{0.48}\\
\bottomrule
\end{tabular}
}
    \caption{\textbf{Performance comparison with various active data selection methods on UTKFace and CINIC-10 datasets.} For both standard and adversarial robust settings, we report the results under three evaluation metrics, including minimax fairness, \ie, highest worst group performance (Worst), performance disparity between the highest and the lowest group (Disp), and group average accuracy (Avg). To ensure the reliability of the experimental results, we conduct three experiments with different random seeds and report the average and standard deviation scores. The result shows that our JIN method achieves the highest standard and robust minimax fairness among all active data selection strategies. Also, our method obtains small performance disparity without incurring average performance loss in most cases.}
    
    \label{tab:main_result}
\end{table*}

%% file: tables/big_tab_new2.tex
\begin{table}[h]
\centering
\setlength\tabcolsep{3pt}
\resizebox{1.0\linewidth}{!}{
    \begin{tabular}{c|ccc|ccc}
    \toprule
    \multicolumn{7}{c}{HAM-10000 Skin Lesion Identification (sensitive groups: \{Male, Female\})}\\
    \midrule
    \multicolumn{1}{c|}{
    \multirow{2}{*}{\begin{tabular}[c]{@{}c@{}}Methods\end{tabular}}}&\multicolumn{3}{c|}{Standard F1-score (\%)}&\multicolumn{3}{c}{Robust F1-score (\%)}\\
    &Worst ($\uparrow$)&Disp ($\downarrow$)&Avg ($\uparrow$)&Worst ($\uparrow$)&Disp ($\downarrow$)&Avg ($\uparrow$)  \\
    
    \midrule
    
    Init. AT& 37.37\std{0.76}&3.62\std{0.51}&39.18\std{0.76}&15.84\std{0.22}&1.92\std{0.49}&16.80\std{0.31} \\ 
    
    \midrule
        RAND&40.20\std{0.24}& 6.02\std{0.93}&43.21\std{0.58}&19.72\std{0.50}&2.26\std{0.20}&20.85\std{0.44}\\
        ENT&44.34\std{1.14}&6.25\std{0.85}&\textbf{47.46\std{1.54}}&20.11\std{0.51}&3.32\std{0.70}&21.78\std{0.20}\\
        CSET&41.89\std{0.65}&3.70\std{0.88}&43.75\std{0.56}&19.86\std{0.91}&2.73\std{0.57}&21.22\std{1.19}\\
        BADGE&43.28\std{1.00}&3.52\std{0.46}&45.04\std{0.83}&20.07\std{0.22}&2.39\std{0.39}&21.27\std{0.29}\\
        \midrule
        G-RAND&36.15\std{1.37}&3.46\std{0.61}&37.88\std{1.67}&16.65\std{0.36}&2.85\std{0.89}&18.07\std{0.66}\\
        MinMax&37.21\std{1.21}&3.59\std{0.86}&39.00\std{1.46}& 16.68\std{0.83}&2.17\std{0.80}&17.77\std{0.72}\\
        OPT&35.53\std{1.45}&4.88\std{1.68}&37.98\std{0.64}&17.32\std{0.38}&\textbf{1.88\std{0.38}}&18.26\std{0.39}\\
        FairAL&43.65\std{0.99}&3.53\std{0.77}&45.42\std{0.68}&19.64\std{0.54}&2.44\std{0.81}&20.86\std{0.83}\\
        \midrule
        \cellcolor{LightCyan}\textbf{JIN}&\cellcolor{LightCyan}\textbf{44.98\std{1.41}}&\cellcolor{LightCyan}\textbf{2.96\std{0.58}}&\cellcolor{LightCyan}46.46\std{1.48}&\cellcolor{LightCyan}\textbf{21.95\std{0.91}}&\cellcolor{LightCyan}2.28\std{0.66}&\cellcolor{LightCyan}\textbf{23.09\std{1.16}}\\
    \bottomrule
    \end{tabular}
}
    \caption{\textbf{Performance Comparison on the HAM-10000 dataset.} Similar to Tab. \ref{tab:main_result}, our JIN method outperforms all baselines in standard and robust minimax fairness. Besides, under severe data imbalance, we observe three group-aware sampling methods (G-RAND, MinMax, OPT) cannot perform well. This issue is specifically discussed in Tab. \ref{tab:ham}.}
    
    \label{tab:main_result2}
    \vspace{-0.5em}
\end{table}

%% file: tables/label_imbl.tex
\begin{table*}[!ht]
\centering

\resizebox{\linewidth}{!}{
    \begin{tabular}{c|cc|cc|cc|cc|cc}
    \toprule
    \multirow{3}{*}{} & \multicolumn{2}{c|}{\textbf{nv}} & \multicolumn{2}{c|}{\textbf{mel}} & \multicolumn{2}{c|}{\textbf{bkl}} & \multicolumn{2}{c|}{\textbf{bcc}} & \multicolumn{2}{c}{\textbf{akiec}}\\
    Methods & \multicolumn{1}{l}{Class}  & F1-score (\%) & \multicolumn{1}{l}{Class} & F1-score (\%) & \multicolumn{1}{l}{Class} & F1-score (\%) & \multicolumn{1}{l}{Class} & F1-score (\%) & \multicolumn{1}{l}{Class} & F1-score (\%)\\
    & Freq (\%) & (STD / Rob) & Freq (\%) & (STD / Rob) & Freq (\%) & (STD / Rob) & Freq (\%) & (STD / Rob) & Freq (\%) & (STD / Rob)\\
    \midrule
        Init. AT&80.11&91.22 / 84.05&4.10&19.04 / \ \ 3.44&\ \ 8.12&41.17 / 10.53&3.30&30.76 / 11.69&2.74&24.06 / ~~0.00\\
    \midrule
        RAND&80.07&92.41 / 86.67&4.12&23.81 / \ \ 4.17&\ \ 8.16&40.90 / 12.12&3.51&43.51 / 13.67&2.54&19.36 / ~~0.00\\
        ENT&69.84&\textbf{93.50 / 86.85}&5.71&39.13 / 11.11&12.36&35.00 / \ \ 9.71&4.65&39.22 / 14.67&5.88&31.82 / 14.26\\
        G-RAND&80.31&92.68 / 86.32&4.04&20.12 / \ \ 2.38&\ \ 8.01&40.87 / \ \ 7.79&3.34&27.61 / 14.78&2.52&25.64 / ~~0.00\\
        FairAL&69.13&92.45 / 85.89&6.13&35.48 / 12.99&12.27&38.46 / 11.41&5.42&44.66 / 15.91&6.13&35.55 / 17.60\\
\cellcolor{LightCyan}\textbf{JIN}&\cellcolor{LightCyan}63.89&\cellcolor{LightCyan}92.50 / 85.82&\cellcolor{LightCyan}6.41&\cellcolor{LightCyan}\textbf{41.67 / 22.22}&\cellcolor{LightCyan}13.58&\cellcolor{LightCyan}\textbf{42.00 / 14.67}&\cellcolor{LightCyan}6.87&\cellcolor{LightCyan}\textbf{45.28 / 16.67}&\cellcolor{LightCyan}6.33&\cellcolor{LightCyan}\textbf{36.73 / 19.67}\\
    \bottomrule
    \end{tabular}
}
\caption{\textbf{Correlation between class frequencies and per-class performance of the worst group on the HAM-10000 dataset.} We report the standard and robust F1-score for the top five high-frequency classes of the worst group. Compared with existing baselines, our JIN method is able to select more samples on less-frequent classes (mel, bkl, bcc, and akiec) and thus achieves significant improvement in standard and robust F1-score. As for the dominant class (nv), though selecting few samples, our method still achieves comparable results with prior approaches.}
\label{tab:ham}
\vspace{-0.5em}
\end{table*}

%% file: tables/computation.tex
\begin{table}[!h]
\centering
\resizebox{0.8\linewidth}{!}{
    \begin{tabular}{c|ccc}
    \toprule
    Methods & UTKFace & CINIC-10 & HAM-10000 \\
    
    \midrule
        Init. AT & 1h 4m 26s & 1h 9m 31s & 1h 22m 7s \\
    \midrule
        ENT & 14s & 45s & 12s \\
        G-RAND &  1m 5s & 2m 17s & 18s  \\
        FairAL &  39m 47s & 2h 21m 29s & 19m 55s  \\
        \cellcolor{LightCyan}\textbf{JIN} &  \cellcolor{LightCyan}10m 29s & \cellcolor{LightCyan}19m 46s & \cellcolor{LightCyan}15m 40s  \\
    \bottomrule
    \end{tabular}
}
\caption{\textbf{Computational costs of different active selection strategies.} Referring to Tab. \ref{tab:main_result} and Tab. \ref{tab:main_result2}, ENT and G-RAND require low computational costs but obtain poor performance, while FairAL has better performance but takes a lot of time. Our JIN method achieves the best standard and robust fairness among all methods with affordable computational cost.}
\label{tab:computation}
\vspace{-0.5em}
\end{table}

%% file: tables/abl.tex
\begin{table}[!ht]
\centering
\resizebox{\linewidth}{!}{
    \begin{tabular}{c|cc|cc}
    \toprule
    \multirow{2}{*}{} & \multicolumn{2}{c|}{{STD. Acc. (\%)}} & \multicolumn{2}{c}{Rob. Acc. (\%)} \\
    & Worst ($\uparrow$) & Avg ($\uparrow$) & Worst ($\uparrow$) & Avg ($\uparrow$) \\
    \midrule
        S &  \textbf{75.18\std{0.47}} & \textbf{75.84\std{0.27}} & 56.53\std{0.08} & 59.30\std{0.11} \\
        R & 72.89\std{0.30} & 74.31\std{0.26} & 56.89\std{0.19} & 59.94\std{0.04} \\
        \cellcolor{LightCyan}\textbf{S+R} &  \cellcolor{LightCyan}75.07\std{0.53} & \cellcolor{LightCyan}75.74\std{0.49} & \cellcolor{LightCyan}\textbf{57.39\std{0.10}}  & \cellcolor{LightCyan}\textbf{60.10\std{0.25}} \\
    \bottomrule
    \end{tabular}
}
\caption{\textbf{Ablation studies on the UTKFace dataset.} The letters S and R stand for standard and robust inconsistency, respectively. Using two inconsistency scores together can achieve the best robust minimax fairness and great standard minimax fairness.}
\label{tab:abl_result}
\vspace{-0.5em}
\end{table}


%% file: tables/worst_group.tex
\begin{table}[!h]
\centering
\resizebox{\linewidth}{!}{
    \begin{tabular}{c|cc|cc}
    \toprule
    \multirow{2}{*}{} & \multicolumn{2}{c|}{{STD. Acc. (\%)}} & \multicolumn{2}{c}{Rob. Acc. (\%)} \\
    & Worst ($\uparrow$) & Avg ($\uparrow$) & Worst ($\uparrow$) & Avg ($\uparrow$) \\
    \midrule
        ENT & 74.10\std{0.79} & 75.33\std{0.56} & 56.94\std{0.64} & \textbf{60.25\std{0.56}} \\
        G-ENT & 68.14\std{0.62} & 70.56\std{0.44} & 54.89\std{0.32} & 58.85\std{0.37} \\
        \cellcolor{LightCyan}\textbf{JIN} &  \cellcolor{LightCyan}\textbf{75.07\std{0.53}} & \cellcolor{LightCyan}\textbf{75.74\std{0.49}} & \cellcolor{LightCyan}\textbf{57.39\std{0.10}}  & \cellcolor{LightCyan}60.10\std{0.25} \\
    \bottomrule
    \end{tabular}
}
\caption{\textbf{A fair comparison in sampling only from the worst-performing group}. G-ENT not only performs worse than our method but even degrades a lot from the original ENT.}
\label{tab:abl_result_main}
\vspace{-0.5em}
\end{table}

%% file: tables/gaware.tex
\begin{table}[!ht]
\centering
\resizebox{\linewidth}{!}{
    \begin{tabular}{c|cc|cc}
    \toprule
    \multirow{2}{*}{} & \multicolumn{2}{c|}{{STD. Acc. (\%)}} & \multicolumn{2}{c}{Rob. Acc. (\%)} \\
    & Worst ($\uparrow$) & Avg ($\uparrow$) & Worst ($\uparrow$) & Avg ($\uparrow$) \\

    \midrule
        Init. AT & 64.80\std{1.79} & 67.46\std{1.39} & 51.48\std{0.41} & 56.42\std{0.23} \\ 
    \midrule
        RAND &  70.86\std{1.46} & 72.83\std{1.01} & 55.40\std{1.36} & 59.52\std{0.81} \\
        ENT & 73.30\std{1.07} & 74.67\std{0.93} & 56.03\std{0.80} & 60.30\std{0.40} \\
        G-RAND &  72.71\std{0.78} & 73.47\std{0.54} & 56.69\std{0.67}  & 59.71\std{0.36} \\
        FairAL &  74.28\std{0.60} & 75.41\std{0.35} & 56.80\std{0.46}  & \textbf{60.68\std{0.35}} \\
        \cellcolor{LightCyan}\textbf{JIN} & \cellcolor{LightCyan}\textbf{75.38\std{0.66}} & \cellcolor{LightCyan}\textbf{75.58\std{0.61}} & \cellcolor{LightCyan}\textbf{57.75\std{0.69}}  & \cellcolor{LightCyan}60.42\std{0.25} \\
    \bottomrule
    \end{tabular}
}
\caption{\textbf{Results on the UTKFace dataset using the ResNet18 model.} The experimental results in Tab. \ref{tab:main_result} and this table confirm that our method achieves advantages over multiple active selection baselines across model architectures.}
\label{tab:gaware}
\vspace{-0.5em}
\end{table}

%% file: tables/multiple_sens.tex
\begin{table}[!h]
\centering
\resizebox{\linewidth}{!}{
    \begin{tabular}{c|cc|cc}
    \toprule
    \multirow{2}{*}{} & \multicolumn{2}{c|}{{STD. Acc. (\%)}} & \multicolumn{2}{c}{Rob. Acc. (\%)} \\
    & Worst ($\uparrow$) & Avg ($\uparrow$) & Worst ($\uparrow$) & Avg ($\uparrow$) \\

    \midrule
        Init. AT & 77.74\std{0.66} & 81.03\std{0.33} & 67.61\std{0.21} & 70.84\std{0.36} \\ 
    \midrule
        RAND & 78.57\std{0.31} & 82.34\std{0.10} & 69.14\std{0.30} & 72.34\std{0.14} \\
        ENT & 80.70\std{0.59} & 84.01\std{0.10} & 69.79\std{0.14} & 72.67\std{0.21} \\
        G-RAND & 81.08\std{0.22} & 82.95\std{0.31} & 70.38\std{0.23} & \textbf{73.39\std{0.29}} \\
        FairAL & 80.41\std{0.60} & 83.78\std{0.32} & 69.78\std{0.30}  & 72.56\std{0.16} \\
        \cellcolor{LightCyan}\textbf{JIN} & \cellcolor{LightCyan}\textbf{82.77\std{0.27}} & \cellcolor{LightCyan}\textbf{84.96\std{0.25}} & \cellcolor{LightCyan}\textbf{70.56\std{0.13}}  & \cellcolor{LightCyan}73.11\std{0.11} \\
    \bottomrule
    \end{tabular}
}
\caption{\textbf{Comparison on the UTKFace gender prediction task under four ethnically sensitive groups.} Under the setting of multiple sensitive groups, our method still outperforms existing baselines in standard and robust minimax fairness.}
\label{tab:multiple_sens}
\vspace{-0.5em}
\end{table}

%% file: 5_conclusion.tex
We introduce a brand new learning framework, Fair Robust Active Learning (FRAL), aimed at eliminating discrimination in safety-critical applications without requiring prohibitive labeling costs. Under this framework, prior data selection strategies suffer from data imbalances and computation burdens. To this end, we propose the JIN method leveraging prediction inconsistencies between standard and robust models as well as benign and adversarial inputs. Validated with diverse datasets and sensitive attributes, our method achieves the highest minimax fairness under standard and adversarial scenarios with limited computations. With our FRAL framework and the JIN method, we anticipate a new era of machine learning research for trustworthy visual applications.

%% file: supp.tex
\section{Implementation Details}

We introduce our computing infrastructure and training details in the supplementary material.

\subsection{Computing Infrastructure}

All experiments are conducted on an 8-core CPU personal computer with an NVIDIA RTX3090 GPU. The computational comparison shown in Tab. 4 in the main paper is evaluated on this machine.

\subsection{Training Details}

The overall framework of our proposed Joint INconsistency method (JIN) is provided in Algo. 1 in the main paper. Here we focus on the more detailed model training process, including the implementation of attack and defense methods as well as a conventional deep neural network pipeline.

\vspace{1em}

\noindent {\textbf{Adversarial Attack and Defense.}} For all experiments, we utilize the python foolbox package \cite{rauber2017foolbox} to achieve PGD-5 white-box adversarial attacks with maximum perturbation range $\ell_{\infty}=4/255$ and step size $\alpha=2/255$. Likewise, leveraging the official TRADES loss \cite{zhang2019trades} implementation\footnote{https://github.com/yaodongyu/TRADES}, we realize the adversarial training with the same perturbation settings as the threat model and set the penalized term as $\beta=6$.

\vspace{1em}

\noindent {\textbf{Deep Neural Network Pipeline.}} In the following, we elaborate on our deep neural network adversarial training and fine-tuning pipeline (corresponding to the ``Adv-TRAIN" and ``Adv-FINETUNE" in Algo. 1 of the main paper). For three datasets, we leverage the SGD optimizer to train our model with an initial learning rate of $\gamma$, a momentum of $\mu$, and a weight decay $\lambda$. The batch size is set to $B$. Initially, we adversarially train the model for $E_0$ epoch. Then, for each active learning iteration, we adversarially fine-tune the model for $E$ epochs. To make the whole training pipeline stable, we utilize the cosine annealing learning rate scheduler with a warm-up stage of initial $E_w$ epochs in both the initialization and fine-tuning stage. 

For the UTKFace dataset, we set $\gamma = 0.1$, $\mu=0.9$, $\lambda=\textrm{2e-4}$, $E_0=100, E=70$, $E_w = 10$, and $B=32$. For the CINIC-10 dadtaset, we set $\gamma = 0.1$, $\mu=0.9$, $\lambda=\textrm{2e-4}$, $E_0=110, E=70$, $E_w = 10$ and $B=64$. For the HAM-10000 dadtaset, we set $\gamma = 0.02$, $\mu=0.9$, $\lambda=\textrm{2e-4}$, $E_0 = E=50$, $E_w = 5$ and $B=32$. Note that for a fair comparison, all models used in our experiments are not pre-trained.